# METHODS AND TOOLS FOR BUILDING THE CATALAN WORDNET

Laura Benítez, Sergi Cervell, Gerard Escudero,
Mònica López, German Rigau and Mariona Taulé.

GTLN-UPC-UB[1]
{laurab, escudero, g.rigau}@lsi.upc.es
{cervell, lopez, mtaule}@lingua.fil.ub.es

**Abstract**

In this paper we introduce the methodology used and the basic phases we followed to develop the Catalan WordNet, and which lexical resources have been employed in its building. This methodology, as well as the tools we made use of, have been thought in a general way so that they could be applied to any other language.

## 1. Introduction

In recent years the research in the Natural Language Processing (NLP) field has proved the need for extensive and complete Lexical Knowledge Bases (LKBs). Acquiring such lexical/semantic structures is a hard problem and has been usually approached by reusing, merging and tuning existing lexical material. While in English the "lexical bottleneck" problem seems to be softened (e.g. WordNet (Miller, 1990), Alvey Lexicon (Grover et al. 1993), COMLEX (Grishman et al. 1994) and so on) there are no available wide range lexicons for NLP for other languages. Manual construction of lexicons is the most reliable technique for obtaining structured lexicons but it is costly and highly time-consuming. This is the reason for many researchers to have focussed on the massive acquisition of lexical knowledge and semantic information from pre-existing structured lexical resources as automatically as possible. The work presented here shows a fast production methodology to build large scale multilingual LKBs from conventional dictionaries.

In this way, the English WordNet developed at Princeton University is consolidating as a *standard de facto* for the lexical-semantic representation of the natural language for English. Considering this English WordNet as a reference, new WordNets in some other languages are being built, such as the ones for Galician, Basque, Spanish and Catalan, for instance. Besides, multilingual links are established automatically through the connections with the English WordNet.

The research we are now presenting is embedded in the Catalan WordNet project, which is being carried out by the *Grup de Tractament de Llenguatge Natural* (GTLN-UPC-UB) of the *Centre de Referència en Enginyeria Lingüística* (CREL)[2].

The aim of this paper is then to show the methodology followed for the building of the Catalan WordNet, as well as which available resources we took advantage of and which ones we had to develop.

The Catalan WordNet is a monolingual lexical knowledge base which initially follows WordNet model developed at Princeton[3], but it is going to be adapted to the EuroWordNet[4] model. In WordNet the concepts are defined as synonymy sets called *synsets* linked one each other through semantic relations (hypernymy, hyponymy, meronymy, antonymy and so on). The grammatical categories represented in WordNet are nouns, verbs, adjectives and adverbs. At present we are working on the nouns and verbs representation.

This paper is organized in three parts. Firstly, it describes the methods which managed the automatic extraction of the noun concepts and their further validation (section 2.1), as well as the methodology for the verbal concepts. Secondly, it explains the developed computing tools. In fact, it describes the design of the *database* (section 3.1) and the *interface* which allows the user to consult and modify the Catalan WordNet (section 3.2). Finally, some conclusions are drawn and we comment the main results, besides further research on the topic (section 4).

## 2. Development Of The Catalan Wordnet

As we previously mentioned, the Catalan Wordnet is a monolingual lexical knowledge base. Although its final aim is the representation of nouns, verbs, adjectives and adverbs, this first stage has been focused in the treatment of noun and verbal concepts, as the Catalan WordNet has followed the same methodology which was applied to Spanish WordNet[5]. Let's see how these concepts are defined in WordNet.

The nouns or nominal concepts are organized hierarchycally which hangs from several nodes, from which the rest of nodes are connected directly or indirectly through the semantic relation of *hyponymy*. The verbs are arranged hierarchically through the relation of

---

*troponymy* and are classified into different semantic fields (verbs of communication, movement, state, change, and so on).

Taking into account the methodology of the EuroWordNet project, a set of *base concepts* has been generated. These are considered the most important concepts which are also common to all the languages involved in this project[6]. This set of *base concepts* are put to use as the starting point for the construction of the Catalan WordNet. To be exact, they are a total of 793 *base noun concepts* and 228 *base verbal concepts* translated by hand. The aim is to try to assure that any of the nodes in the hierarchy is connected, directly or indirectly, to one of these *base concepts*.

Next, we will explain the methodology employed in extracting, generating and validating the noun and verbal concepts which are not *base concepts*.

## 2.1. Noun Concepts

The generation of noun concepts has been done automatically from machine readeable versions of bilingual dictionaries following (Atserias et al. 1997). In this project the Diccionari Anglès-Català Català-Anglès de la Enciclopèdia Catalana (DEC, 1996) has been used. All the possible pairs of *English word - Catalan word* are taken out of this concrete dictionary, so that they could be connected to the synsets of WordNet (Benítez et al. a, 1998). Once we obtained these pairs, some methods are applied to generate this set of three elements, *Catalan word - English word - synset* and to disambiguate them in relation to Wordnet. Such methods are the *class methods* (Atserias et al. 1997). Their application consists in doing a complete and discrete partition of the set of pairs *Catalan word - English word* into 4 disjoint subsets:

a) In the first of these subsets there are the Catalan Words (CW) which have only one translation into a unique English Word (EW) and, besides, this EW can only be translated into that CW.

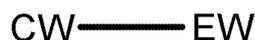

**Figure 1: Criterion 1.**

b) In the second one there are the CWs which have more than one translation into English and each one of these EWs can only be translated into that unique CW.

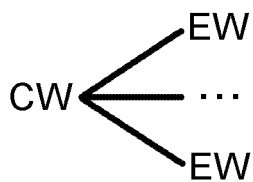

**Figure 2: Criterion 2.**

c) In the third subset we find the EWs which have only one translation into a unique CW and, besides, this CW can only be translated into that EW.

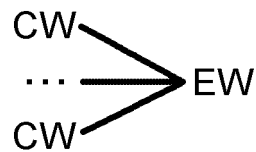

**Figure 3: Criterion 3.**

d) Finally, in the fourth one there are the CWs which have more than one translation into EWs and every one of these EWs is translated into more than one CW.

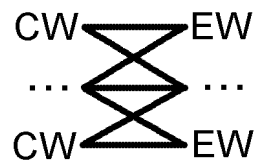

**Figure 4: Criterion 4.**

Subsequently, the set of *English word - synset* pairs is divided into two subsets: the one of *monosemic* pairs and that of *polysemic* ones. Then these two subsets of *English word - synset* are joined with those four *Catalan word - English word* subsets. As a result we obtain 8 new groups of sets of three elements, *Catalan word - English word - synset,* where the *English word* is the same in every two joined subsets (Benítez et al. b, 1998)[7].

Up to here, we have explained the application of monosemic and polysemic criteria of the *class methods*. We have also used the *variant criterion*, an hybrid criterion from the *class methods*, which consists in linking *Catalan words* to *synsets*. The link is established when the synset contains a set of English words and two or more of these English words have as a translation the same Catalan word.

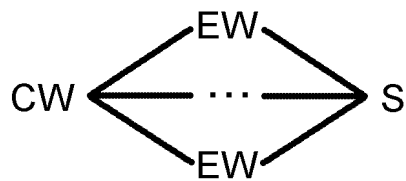

**Figure 5: Variant criterion**

From each of those obtained 9 sets, a representative random sample has been extracted and validated by hand, in order to calculate every confidence score. The reliability obtained from the sample is assigned to the whole set. We have validated 1515 out of 49069 *Catalan word - English word - synset* links in this thorough way.

---

[6] The languages involved in the EuroWordNet project are English, Spanish, Italian and Dutch. The base concepts has been extracted by means of the evaluation of the frequencies from textual corpora and genus from Machine Readeable Dictionaries.

[7] In this paper we can see the practical application of the *class methods* explained in the paper *Combining Multiple Methods for the Automatic Construction of Multilingual WordNets* (Atserias et al. 1997), as well as an implementation with Unix tools.

The subsets with a confidence score over 85 % have been added to the noun concepts, in order to increase the Catalan WordNet.

| Criteria | #links | #synsets | #words | %ok |
|---|---|---|---|---|
| mono1 | 1226 | 1212 | 1221 | 95.9 |
| mono2 | 419 | 337 | 258 | 97.6 |
| mono3 | 448 | 208 | 396 | 93.3 |
| mono4 | 3012 | 1532 | 2178 | 94 |
| poly1 | 2298 | 2244 | 864 | 90.4 |
| poly2 | 568 | 519 | 158 | 77.9 |
| poly3 | 1125 | 477 | 357 | 71.7 |
| poly4 | 37714 | 9151 | 4266 | 54.5 |
| variant | 2259 | 1517 | 1516 | 96 |

**Table 1: Results of class methods.**

At the present, the represented noun synsets are 7509, which means 6961 noun forms and 10705 links.

## 2.2. Verbal Concepts

The methodology employed in representing the verbal concepts has been slightly different from the one in the noun concepts. Bearing in mind the accented polysemy we can find in the verbs and considering the lower number of verbs too, a complete manual building was chosen.

On one side and thanks to the Pirápides[8] project, where some of the members of our team took part, a list of English verbs was available, being these ones classified semantically, as in the proposal of Beth Levin[9] (Levin, 1993), with their translations into Spanish and Catalan. On the other side, the professor Bonnie Dorr at the Maryland University kindly provided us with the correspondence *English verb - Levin class - synset* (Dorr et al. 1997). From these two lists the link *Catalan verb - synset* was generated automatically before we validated every link carefully.

The total amount of verbal concepts is 3244, which means a whole of 3359 verbal forms and 9121 links.

## 3. Software

The Catalan WordNet is being modifying and updating constantly. What is more, all these changes have to be done by any of the members of our group at the same time. This is why we need a database which assures the integrity of the information contained and this simultaneous access, in short, its consistency. To make the user work easier, a friendly interface has been created. Firstly, this one allows the user to search and manipulate the data easily and effectively. The user also can consult multilingual data and have an intuitive view of the hierarchy of the synsets. Secondly, some lexical data (mainly monolingual and bilingual dictionaries) are available.

### 3.1. Database

To minimize the developing expenses and to let the free access to the data, a freeware database, the *Mini SQL* de *Hughes Technologies*[10] (Mini SQL) was used. This one can be installed in any Unix system (including Linux) and offers Application Program Interfaces (APIs) for the main computing languages.

The main objectives in the relational database design are the storage of the general information of the different WordNets, as well as to keep all this information in a well defined structure (Benítez et al. c, 1998). At the moment our design includes the complete loading of WordNet 1.5, which means the following:

a) The *synsets* and their related information (the gloss, the semantic field and the total amount of direct and indirect hyponyms),
b) the different meanings of every word and its connection to the *synset*, and
c) the semantic relations stored as connections for *synsets* (hyperonymy, hyponymy, antonymy, meronymy, holonymy, attribute, cause and entailment).

But for the rest of WordNets we kept:
a) the meanings of every word with its connection to the *synset* in WordNet 1.5 besides the reliability obtained through the methods explained in section 2, and
b) the gloss related to the *synset*.

Right now the semantic relations in WordNet 1.5 form the skeleton of the hierarchy of all the WordNets in the system. The multilingual connection of the meanings of the words is also done through the *synsets*. The present loaded WordNets are fragments of the Spanish, Basque and Catalan ones.

Integrating new WordNets in new languages can be easily done, and the structure of the database allows to develop exporting *software* easily, too. Actually, a prototype for exporting monolingual Wordnets capable for the tools[11] in EuroWordNet (Novell, 1996-1998) has already been implemented.

### 3.2. The Interface

In order to build the Catalan WordNet or any other monolingual WordNet it is necessary the creation of an accessing interface to the database and to the different available lexical resources. Due to the physical distance between the different members of the project who handle the data and to allow any user to consult it openly, the interface has been developed to work through the Web (Benítez et al. c,d, 1998).

The operations the interface offers are the following:

- **Consult Multilingual WordNet**. Its main goal is to aid the checking of errors and gaps in the hierarchic arrangement of the different WordNets. To make this possible the interface shows the path through the semantic hierarchy of one of the WordNets, for any semantic relation, having as a starting point a synset, a sense or a word of one of the available languages.
- **Edit Multilingual WordNet**. Obviously its basic target is the edition of the words and glosses of all of the WordNets, except for those of WordNet 1.5. For every verb it is also possible to edit its corresponding Levin semantic classes (see section 2.2).
- **Consult Lexical Resources.** It allows to consult the different bilingual and monolingual dictionaries. This information is not storaged in the database, but into text files. At the moment these lexical resources availables for Catalan are the <u>Diccionari general de la llengua catalana</u> (Fabra, 1984), and the <u>Diccionari bàsic català-anglès anglès-català de l'Enciclopèdia Catalana</u> (DEC, 1996).
- **Edit Statistics.** This option shows an historic list of the changes and updates done through the interface.

---

[8] Pirápides is a project in which the *Universitat de Barcelona*, *Universitat Autònoma de Barcelona*, *Universitat de Lleida* and *Universidad de San Sebastián* take part. Its main aim is to create lexical corpora for natural language processing and dictionaries of use.

[9] The verbal classification proposed by Beth Levin is based on the hypothesis which states that the syntactic behaviour is determined by the meaning of a verb, therefore, it is useful to study the diathesis alternations to classify verbs semantically.

[10] The address of this company is "*http://www.Hughes.com.au*".
[11] Polaris 1.3. of Novell Inc.

automatic and complementary techniques for linking Catalan words collected from bilingual MRDs (for nouns) and lexicons (for verbs) to English WordNet synsets. As the Catalan WordNet produced in this way is neither error free nor complete, we implemented a Web interface allowing manual interaction (that is, consulting and editing) with the multilingual WordNet. Currently, due to the smaller coverage we achieved by mapping the Catalan/English bilingual MRD (compared to the size of WordNet1.5), we are planning to join Catalan taxonomies acquired automatically from a monolingual MRD (Rigau et al. 1997). The results show that the construction of multilingual LKBs can be performed quickly using a mixed methodology. That is, building a multilingual LKB derived from bilingual MRDs by using automatic procedures, and then performing, in a second stage, a manual refinement of the acquired data.

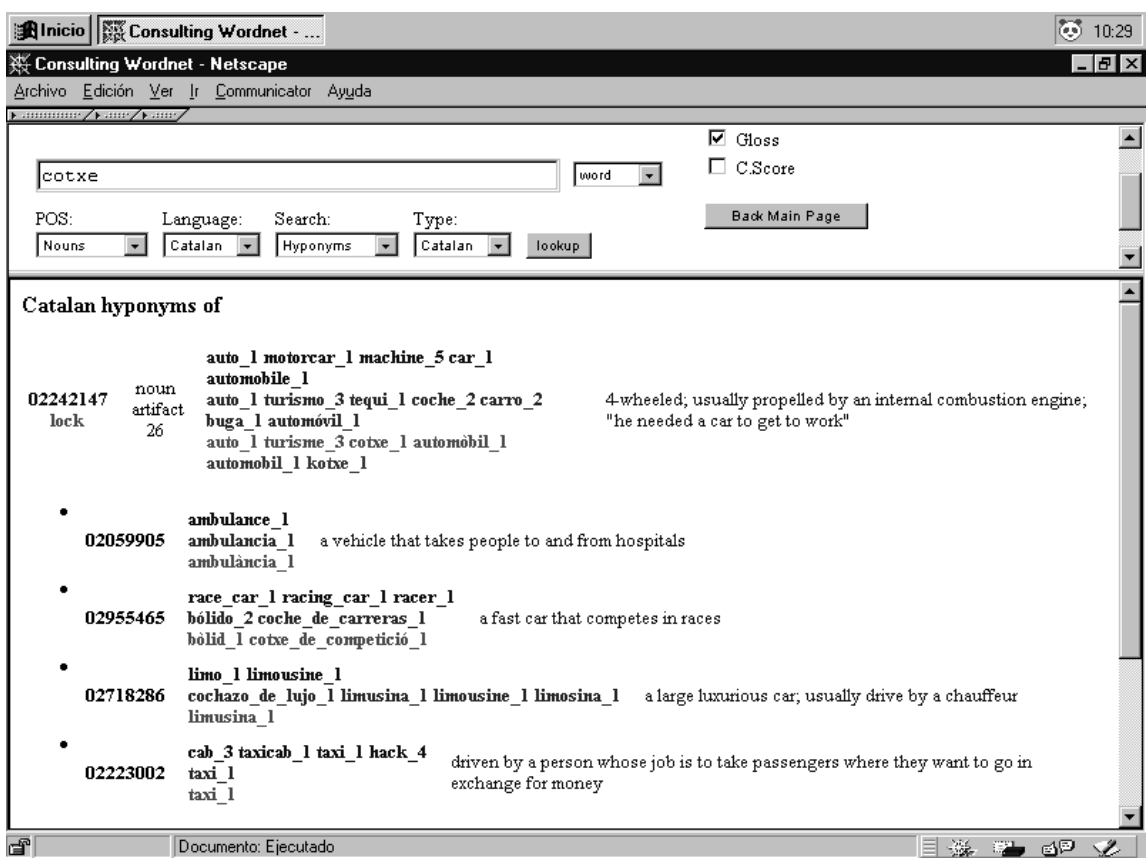

**Figure 6: Example of Consult Multilingual WordNet.**

In the implementation of the interface we made use of some programming languages such as Perl, HTML and JavaScript (Aronson & Lowery, 1997; CGI; Goodman, 1996; Muelver, 1996; Wall et al., 1996].

## 4. Conclusions And Future Perspectives

The first steps that have been taken towards producing a preliminary version of the Catalan WordNet has been described. This work proposes the automatic construction of the core of a multilingual Lexical Knowledge Base from preexisting lexical resources. First, following (Atserias et al 1997), we explain a set of